\documentclass[acmsmall,screen,nonacm]{acmart}
\acmJournal{PACMPL}
\acmNumber{PLDI} 
\acmYear{2024}
\acmMonth{1}
\acmDOI{} 
\startPage{1}

\usepackage{algorithmic}
\usepackage{graphicx}
\usepackage{textcomp}
\usepackage{listings}
\usepackage{xcolor}
\usepackage{url}
\usepackage{multirow}
\definecolor{light-gray}{gray}{0.80}

\usepackage{tikz}
\usetikzlibrary{tikzmark}

\newcommand{\mytikzmark}[1]{%
  \tikz[overlay,remember picture,baseline] \coordinate (#1) at (0,-0.05) {};}

\newcommand{\highlight}[2]{%
  \draw[gray,line width=8pt,opacity=0.5]%
    ([yshift=4pt]#1) -- ([yshift=4pt]#2);%
}

\newcommand{\fsd}{\texttt{FSD}}
\newcommand{\invd}{\texttt{InvD}}

\usepackage{listings}
\usepackage{xcolor}

\begin{document}

\title{Transfer Attacks and Defenses for Large Language Models on Coding Tasks}

\author{Chi Zhang}
\email{chiz5@andrew.cmu.edu}
\affiliation{%
  \institution{Carnegie Mellon University}
  \country{USA}
}

\author{Zifan Wang}
\email{zifan@safe.ai}
\affiliation{%
  \institution{Center for AI Safety}
  \country{USA}
}

\author{Ravi Mangal}
\email{rmangal@andrew.cmu.edu}
\affiliation{%
  \institution{Carnegie Mellon University}
  \country{USA}
}

\author{Matt Fredrikson}
\email{mfredrik@cmu.edu}
\affiliation{%
  \institution{Carnegie Mellon University}
  \country{USA}
}

\author{Limin Jia}
\email{liminjia@andrew.cmu.edu}
\affiliation{%
  \institution{Carnegie Mellon University}
  \country{USA}
}

\author{Corina P\u{a}s\u{a}reanu}
\email{pcorina@andrew.cmu.edu}
\affiliation{%
  \institution{Carnegie Mellon University}
  \country{USA}
}

\begin{abstract}
  Modern large language models (LLMs), such as ChatGPT, have demonstrated impressive capabilities for coding tasks including writing and reasoning about code. 
They improve upon previous neural network models of code, such as code2seq or seq2seq, that already demonstrated competitive results when performing tasks such as code summarization and identifying code vulnerabilities. 
However, these previous code models were shown vulnerable to adversarial examples, i.e. small syntactic perturbations that do not change the program's semantics, such as the inclusion of ``dead code'' through false conditions or the addition of inconsequential print statements, designed to ``fool'' the models.  
LLMs can also be vulnerable to the same adversarial perturbations but a detailed study on this concern has been lacking so far. 
In this paper we aim to investigate the effect of adversarial perturbations on coding tasks with LLMs. 
In particular, we study the transferability of adversarial examples, generated through white-box attacks on smaller code models, to LLMs.
Furthermore, to make the LLMs more robust against such adversaries without incurring the cost of retraining, we propose prompt-based defenses that involve modifying the prompt to include additional information such as examples of adversarially perturbed code and explicit instructions for reversing adversarial perturbations. 
 Our experiments show that adversarial examples obtained with a smaller code model are indeed transferable, weakening the LLMs' performance. 
The proposed defenses show promise in improving the model's resilience, paving the way to more robust defensive solutions for LLMs in code-related applications.

\end{abstract}


\maketitle

\section{Introduction}
\label{sec:intro}

Modern large language models (LLMs), such as ChatGPT\footnote{\url{https://chat.openai.com}}, have fundamentally changed the landscape of computational tasks. These LLMs, characterized by their ability to understand and generate human-like text across a wide range of topics, have demonstrated remarkable capabilities in coding tasks. They perform roles like writing code~\cite{DBLP:journals/corr/abs-2108-07732,DBLP:journals/corr/abs-2107-03374,doi:10.1126/science.abq1158,liu2023is} and reasoning about its functionality~\cite{pmlr-v202-pei23a,xia2023automated}, and in doing so, they go beyond the capabilities of existing neural network models that have already shown promise in tasks such as code summarization~\cite{allamanis2016convolutional,alon2018codeseq,alon2019code2vec} and identifying code vulnerabilities~\cite{DeepBugs,allamanis2018learning,vasic2018neural,Hoppity}.

While these LLMs offer clear advancements, they also inherit vulnerabilities from their predecessors. One notable weakness is susceptibility to adversarial examples, a form of machine learning attack that has been well-documented in previous research~\cite{szegedy2014intriguing,GoodfellowSS14,UniversalAttack23}. Adversarial examples consist of subtly altered inputs designed to mislead machine learning models without changing the underlying meaning of the data. In code-related tasks, these could involve the insertion of {\em dead code} via false conditions or the addition of inconsequential print statements, both of which are aimed at leading the model astray \cite{srikant2021generating, yefet2020adversarial, gao2023discrete, allamanis2018learning}.

Despite the increased utility and application of LLMs in coding tasks, research that systematically evaluates their vulnerability to these types of adversarial attacks is lacking, prompting the need for our study. We aim to evaluate LLMs' resilience, or lack thereof, to adversarial manipulations—particularly those attacks that have proven effective against smaller, specialized code models, and are therefore cheap to compute. 

Through experiments, we show that adversarial examples obtained with a state-of-the-art attack technique for a smaller code model (seq2seq~\cite{sutskever2014sequence,srikant2021generating}) are indeed transferable to multiple LLMs (GPT-3.5 and GPT-4 from OpenAI~\cite{openai2023gpt4}, Claude-Instant-1 and Claude-2 from Anthropic~\cite{anthropic}
), weakening the analyzed LLMs’ performance.

To defend against such attacks and improve the resistance of the models against adversarial examples, we discuss post-hoc approaches, i.e., ones that do require retraining or fine-tuning the model. This is necessitated by the black-box nature of commercial LLMs. In particular, we propose novel {\em self-defense} strategies that leverage LLMs's own advanced capabilities of performing in-context learning~\cite{brown2020language} and understanding human instructions as well as code. The self-defense strategies involve modifying a manually crafted prompt to include additional information such as examples of adversarially perturbed code and explicit instructions for reversing adversarial perturbations.
Our experimental evaluation
of the proposed defenses indicates promise in improving the model’s resilience, paving the way
to more robust defensive solutions for LLMs in code-related applications.

We also present \emph{meta-prompting} for leveraging the LLMs themselves to generate the self-defense prompts. The performance of LLMs can be very sensitive to the prompts used but, at the moment, prompt design is an empirical process lacking sound principles. We meta-prompt an LLM with examples of perturbed and corresponding unperturbed snippets, and ask the model to synthesize a prompt that can be used to unperturb the code using an LLM. Our experiments show that self-defense prompts generated via meta-prompting can be much more effective than the prompts that are manually crafted.

We summarize our contributions as follows: (1) We study the transferability of code attacks obtained based on a small code model to five commercial and open-source state-of-the-art LLMs. Through experimental results we find that all studied LLMs are susceptible to such attacks, with an attack success rate (ASR) of at least 21\% observed on GPT-4.  
(2) We propose self-defense techniques against transfer attacks that leverage LLMs' own capabilities of performing in-context learning and understanding human instructions. This defense reduces the effects of the attack, achieving e.g., an ASR of 14\% for GPT-4 and reduced rates for other models.
(3) We also propose a meta-prompting technique that uses an LLM to generate its own defense prompt, further reducing the attack effectiveness, e.g., obtaining an ASR of 4\% for GPT-4.
(4) A secondary contribution of this work is a set of techniques aimed to customize LLMs for code summarization tasks.
\section{Background}
\label{sec:background}

\subsection{Code Models: Modern LLMs vs. Previous Code Models}

In this paper we make the distinction between modern, state-of-the-art LLMs, and previous, smaller specialized code models (which we denote as pre-LLM), for which known, practical attack methods exist.
We distinguish them in terms of capabilities and sheer size. Modern, state-of-the-art LLMs, such as the proprietary GPT-4 \cite{openai2023gpt4} and open-source Llama-2~\cite{touvron2023llama}, generally have a transformer~\cite{NIPS2017_3f5ee243} architecture, capable of following instructions, and are trained on general-purpose data, not only code. Smaller specialized models, such as code2vec~\cite{alon2019code2vec}, code2seq~\cite{alon2018codeseq}, or seq2seq~\cite{srikant2021generating, alon2018codeseq}, have various types of architectures and have been trained specifically for coding tasks.
These models are much smaller than the modern LLMs (e.g., millions vs. billions of parameters), and are, therefore, easier to analyze. While most of existing LLMs are decoder-only, there are also encoder-based powerful models, such as BERT~\cite{devlin2019bert} 
and code-specific CodeBERT~\cite{feng2020codebert}, that, however, are still much smaller than the general purpose LLMs, and as such we would include them in the second category. 

We aim to evaluate whether the attacks computed for the much smaller models, which are relatively cheap to compute, transfer to the much larger, general-purpose modern LLMs. Of course, an interesting question is how to compute attacks that are optimized directly on the LLMs. However, such attacks are very expensive in practice but we hope to explore this direction in future work.

\subsection{Attacks on Pre-LLM Code Models}
\label{attack_seq2seq}
An adversarial example is a seemingly benign input with well-crafted tiny perturbations that maliciously cause a machine learning (ML) model to make a mistake, e.g. a wrong prediction. Adversarial perturbations can be very small (imperceptible to the human eye) in the case of perception models \cite{GoodfellowSS14,szegedy2014intriguing}. Code models are susceptible to adversarial perturbations as well. Such attacks can have adverse consequences in settings such as security and compliance automation~\cite{srikant2021generating}. An adversary can perturb malign programs in such a way that a classifier predicts them as \texttt{benign} (while they are actually malicious) or can make changes to pass off open-source code in a proprietary code base.

In this section, we review the attack proposed by \citet{srikant2021generating} and reuse it in our studies. 
The attack leverages program obfuscations, which have traditionally been used 
to avoid attempts at reverse
engineering programs, as adversarial perturbations. These perturbations modify
programs in ways that do not alter their functionality (i.e., they are semantics preserving) yet they deceive an ML model when making a decision. 

Specifically, \citet{srikant2021generating} describes a way to generate adversarial programs for a pre-LLM code model, which is trained to solve the code summarization task, i.e., given the code corresponding to the body of a function, the model predicts the function name from a pre-defined set of names. These adversarial examples are generated by solving a constrained combinatorial optimization problem. 
Their method employs two kinds of semantics-preserving changes, i.e. \textit{replace} and \textit{insert}, to the original benign program in order to transform it into an adversarial example. Namely, \textit{replace} changes include actions such as \texttt{renaming local variables}, \texttt{renaming function parameters}, \texttt{renaming object fields}, and \texttt{replacing boolean literals}.
The \textit{insert} changes involve steps such as \texttt{inserting print statements} and \texttt{adding dead code}. The attacker is assumed to have white-box access to the code model for the purpose of generating the adversarial inputs.

The authors address two key problems:
\begin{itemize}
    \item \textit{Site-Selection Problem}: This problem posits the question: given \( n \) sites in a program, if we are permitted to select at most \( k \) sites, which subset of \( k \) sites would exert the most significant impact on the performance of the downstream model?
    \item \textit{Site-Perturbation Problem}: Once the \( k \) sites have been selected, the question arises as to which tokens should be inserted or replaced at these chosen locations.
\end{itemize}

Generating an adversarial program is formulated as an optimization problem: 
\begin{equation}
\label{eqn:adv_attack}
\begin{aligned}
& \underset{\mathbf{z},\mathbf{u}}{\text{minimize}}
& & l_{\text{attack}}((\mathbf{1}-\mathbf{z}) \cdot P + \mathbf{z} \cdot \mathbf{u}; P, \boldsymbol{\theta}) \\
& \text{subject to}
& & \mathbf{1}^T \cdot \mathbf{z} \leq k, \\
& & & \mathbf{z} \in \{0,1\}^n, \\
& & & \mathbf{1}^T \cdot \mathbf{u}_i = 1, \; \mathbf{u}_i \in \{0,1\}^{|\Omega|}, \; \forall i
\end{aligned}
\end{equation}

Here \(l_{\text{attack}}( )\) represents the loss of the attack algorithm;
the authors use cross-entropy loss 
in an untargeted setting. \(P\) represents a benign program,
 composed of a sequence of \(n\) tokens \(\{P_i\}^n_{i=1}\); 
each \(\{P_i\}\in \{0,1\}^{|\Omega|}\) is considered a one-hot vector of length \(|\Omega|\), where \(\Omega\) is a vocabulary of tokens.
\(\mathbf{z}\) is a vector of boolean variables indicating whether a site is selected for perturbation or not. 
\(\mathbf{u}\) is a one-hot vector encoding the selection of a token from \(\Omega\), designated to be the insert/replace token for a selected transformation at a chosen site. 
\(\boldsymbol{\theta}\) denotes the parameters of the given code model. %

Due to the discrete nature of the problem, the loss landscape of the optimization problem in Equation~\ref{eqn:adv_attack} is not smooth. \citet{srikant2021generating} further propose to use randomized smoothing~\cite{duchi2012randomized}, that convolves the original loss landscape with a smooth function, to smoothen the overall loss landscape and ease the optimization problem, which improved their attack success rate. 
Through these methodologies, adversarial examples were generated for two code models based on the seq2seq~\cite{sutskever2014sequence} architecture. 
The underlying code models were  trained on a dataset with 150K Python samples \cite{10.1145/3022671.2984041} and one with 700K Java samples \cite{alon2018codeseq} respectively. In the rest of this paper, we only consider the model trained on Python samples.
The authors 
generate perturbations only for the samples that the model had originally correctly predicted
and evaluate the attack success rate. 
We utilize these adversarial examples in our experiments to evaluate the transferability of the attack from a pre-LLM to LLM code model; furthermore, we use these examples to evaluate the effectiveness of our prompt-based defenses.

\begin{lstlisting}[language=Python, caption={Original Python code example. The snippet is missing indentations since the baseline seq2seq model consumes code in this form.}, label=code:original, float=t]
( self application name = python gntp notifications = [ ] default notifications = none application icon = none hostname = localhost password = none port = 23053 ) : self . application name = application name self . notifications = list ( notifications ) if default notifications : self . default notifications = list ( default notifications ) else : self . default notifications = self . notifications self . application icon = application icon self . password = password self . hostname = hostname self . port = int ( port )
 \end{lstlisting}

\begin{lstlisting}[language=Python, caption={Adversarial Python code example. The sections highlighted in gray are the adversarial perturbations added to the code by the attack.}, label=code:Adv, float=t]
( self application name = python gntp notifications = [ ] default notifications = none application icon = none hostname = localhost password = none port = 23053 ) :    self . application name = application name    self . notifications = list ( notifications )    if default notifications : @\mytikzmark{hl1Start}@if false : traverse = 1@\mytikzmark{hl1End}@   self . default notifications = list ( default notifications )    else : self . default notifications = self . notifications    self . application icon = application icon  @\mytikzmark{hl2Start}@print ( pspace )@\mytikzmark{hl2End}@  self . password = password    self . hostname = hostname @\mytikzmark{hl3Start}@if false : validity = 1@\mytikzmark{hl3End}@   self . port = int ( port ) @\mytikzmark{hl4Start}@if false : sl = 1  if false : evoked = 1@\mytikzmark{hl4End}@
 \end{lstlisting}

 \begin{tikzpicture}[remember picture, overlay]
        \highlight{hl1Start}{hl1End}
        \highlight{hl2Start}{hl2End}
        \highlight{hl3Start}{hl3End}
        \highlight{hl4Start}{hl4End}
    \end{tikzpicture}  

Listing~\ref{code:original} shows an original Python code snippet from the clean dataset, while Listing~\ref{code:Adv} shows the corresponding adversarial example generated using the attack algorithm of \citet{srikant2021generating}.
This code snippet corresponds to the \texttt{\_\_init\_\_} function of a class used for sending messages using a transport protocol. 
In this case, we see that the adversarially perturbed version has a print statement and multiple branches with false condition inserted.
\section{Using Large Language Models for Code Summarization}
\label{sec:LLM-code-summary}

\begin{figure*}[t]
    \centering
    \includegraphics[width=0.9\textwidth]{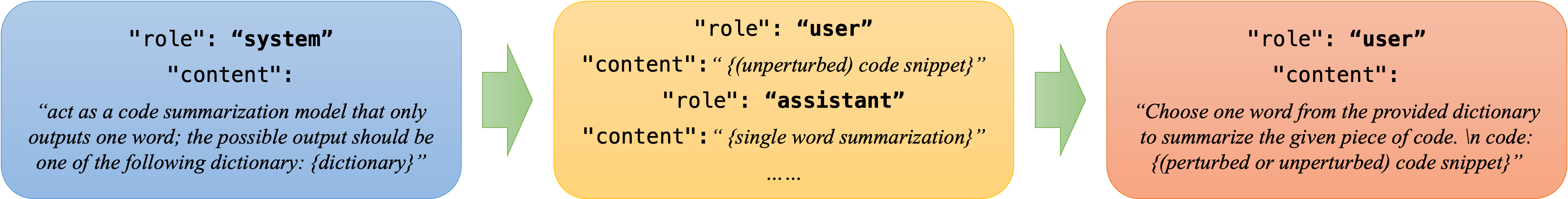}
    \caption{LLM prompt for the code summarization task. The sections of the prompt represented by $\{\cdot\}$ are substituted with the corresponding values in the actual prompt. 
    Prompts are structured as a conversation comprising a sequence of messages. Each message consists of a \texttt{role} identifying the speaker and \texttt{content} of the speaker's words. In this case, the prompt starts (leftmost box) with a \textbf{system} message that helps the model produce outputs in the desired format. Next (middle box), a series of messages between \textbf{user} and \textbf{assistant} with unperturbed code snippets and corresponding function names serve as few-shot examples of the desired behavior. Note that \textbf{assistant} refers to the model. Finally, the prompt ends (rightmost box) with the actual code snippet to be summarized.}
    \label{fig:summary}
\end{figure*}

This section introduces our use of modern LLMs to do code summarization before we delve into the implementation of the transfer attack to the task in the following sections.

The LLMs of interest in this work are \emph{generative} models, taking a string of text, which is often referred to as a \emph{prompt}, and generating a completion to the input prompt. Prompts often include instructions, e.g. \textit{Outline a plan to organize a party at my home}, and LLMs are tuned  to follow the instruction in the prompts to generate corresponding completions~\cite{ouyang2022training}. To use generative LLMs for code summarization, we prompt the LLMs to choose a single class from a list of potential classes that is most likely to describe the functionality of the program in the prompt.
Figure~\ref{fig:summary} showcases our prompt templates for querying GPT-3.5 and GPT-4 models.

\subsection{Preliminary Approach}

As is described in Section~\ref{sec:background}, we take clean and adversarial data from~\citet{srikant2021generating} to evaluate the robustness of other LLMs. 
As our first step towards ensuring that the model's outputs are aligned with the desired output format (i.e., a single word function name from a pre-defined set of function names), we experiment with integrating the \emph{dictionary}\footnote{We refer to a set of potential function names that the code summarization classifier can choose from as the \emph{dictionary}.} of potential function names into the prompt. Our initial prompt takes the following form where \textit{$\{$dictionary$\}$} is substituted by all possible function names and \textit{$\{$code$\}$} by the function body under consideration (for instance, either Listing~\ref{code:original} or ~\ref{code:Adv}):

\noindent\fbox{%
\parbox{\textwidth}{
\centering 
\textit{Use one word from the set $\{$dictionary$\}$ to describe the following piece of code: $\{$code$\}$. Don’t provide any other description. The reply form should be "word".}}}

Initially, we attempt to use a comprehensive dictionary containing all possible target function names (approximately 14,000 unique names for the dataset under consideration). 
However, this approach is problematic as the total length of the prompts exceeds the available context length of the model (4,096 tokens for GPT-3.5 which translates roughly to 3000 words), necessitating a modification in our strategy.
To overcome the constraints, we revise our strategy by tailoring the dictionary to only contain 500 function names including the correct function name for the code being summarized. All the names, apart from the correct one, are randomly chosen from the set of all function names. 
We acknowledge that utilizing a smaller and more focused dictionary inherently simplifies the code summarization task. 
However, the goal of our experiments is to evaluate the susceptibility of LLMs to adversarial attacks rather than the inherent accuracy of the LLMs for code summarization. Any change that aids model accuracy, has the effect of reducing the model susceptibility to adversarial attacks. Therefore, the attack success rates we report in our evaluations should be considered a lower bound, with the actual situation likely to be worse.

Although the use of the dictionary in the prompt is helpful, the LLM demonstrated variability in adhering to the desired format, sporadically generating more extensive sentences, or indulging in a detailed analysis of each line of code, thus deviating from the expected succinct response format.

\subsection{Conforming with Output Format via \texttt{System} Role}
\label{subsec:role-sys}

Faced with the challenge of the models not adhering to the required output format (i.e., single-word function names from a pre-determined set of names), we turned to the functionality of \emph{roles} provided by LLMs. Roles are special words or tokens in the prompt that LLMs are trained to recognize as indicators of who is speaking, and they can have a large effect on how the model interprets the prompt.
LLMs typically recognize at least three roles, namely, \texttt{System}, \texttt{User}, or \texttt{Assistant}, with each having a unique impact on shaping the conversation.

\begin{itemize}
    \item \texttt{System}: This role, while optional, is crucial for setting up the model's behavior by providing high-level instructions or context. Instructions provided under this role tend to be treated as a higher privilege and hence, ones that need to be strongly followed.
    \item \texttt{User}: This role represents input prompts from the end-user or an application. This is the typical role used when providing prompts to the model in an interactive session.
    \item \texttt{Assistant}: This role signifies responses from the model. When engaged in a dialogue with the model, the entire history of the conversation is included in each prompt. In this history, the previous model responses are designated as being generated under this role, and therefore, it is essential for maintaining the continuity of dialogue.
\end{itemize}

To address the challenge of LLMs outputs not conforming to the desired output format, we strategically leveraged the \texttt{System} role to influence the model’s behavior. 
By setting the following explicit instruction through this role,
\fbox{%
\textit{act as a code summarization model that only outputs one word}},
we achieved greater conformance with the desired output format compared with the use of the default \texttt{User} role.

We also explored the effect of providing the dictionary of function names to the model through the \texttt{System} role.
Our experiments revealed that incorporating the dictionary through the \texttt{System} role yielded more structurally well-formed results than doing so through the \texttt{User} role. 
Based on this observation, we established the final prompt for the \texttt{System} role as (leftmost box in Figure~\ref{fig:summary}):

\noindent\fbox{%
\parbox{\textwidth}{
\centering
\textit{act as a code summarization model that only outputs one word; the possible output should be one of the following dictionary: \{dictionary\}}.}}

\subsection{Improving Accuracy via Few-shot Prompting}
\label{subsec:few-shot}

Having designed a prompt that helps the model produce outputs in the format needed for the code summarization task, our next step was to figure out prompting strategies to improve the model's accuracy.
For this, we employed prompt-based few-shot learning.
Few-shot learning~\cite{wang2020generalizing} refers to the notion of modifying the behavior of a machine learning model using very few examples, and this is typically achieved by fine-tuning the model parameters using a gradient descent based algorithm.
However, a striking observation about the behavior of LLMs has been that they are able to function as few-shot learners without any updates to their parameters~\cite{brown2020language}.
They can learn to perform new tasks by simply being exposed to a limited number of examples in the model prompt, negating the need for large datasets or extensive fine-tuning.

In our experiment, we utilized few-shot learning by introducing a series of messages between the \texttt{User} and the \texttt{Assistant} as examples in the prompt (see middle box of Figure~\ref{fig:summary}). Each of these \texttt{User} messages included unperturbed code corresponding to the body of a function (for instance, Listing~\ref{code:original}) while the \texttt{Assistant} messages included the associated function names.
These examples were designed to prime the model for the task of code summarization. 
Employing this few-shot learning approach improved adaptability and accuracy in performing the code summarization task, enhancing the model's ability to generate accurate code summaries.

\subsection{Assessing Model Uncertainty via Self-reported Probabilities}
\label{subsec:confidence}

We experimented with the idea of asking the model to output its confidence along with the predicted function name. The confidence is a measure of the model's subjective probability of its prediction being correct. If confidence predictions of the model are well-calibrated, i.e., probability associated with the predicted function name
reflects its ground truth correctness likelihood, then the predicted confidence can be used to decide if the predicted function name is likely to be a misclassification or not. This can be particularly useful for detecting adversarial perturbations. We also hoped that having the model output its confidence would aid in improving its accuracy. Recent empirical results suggesting that LLMs are indeed able to output well-calibrated confidence scores for various tasks~\cite{kadavath2022language,tian2023just} served as initial evidence for the validity of this approach.
Accordingly, to evaluate this strategy, we constructed the following prompt:

\noindent\fbox{%
\parbox{\textwidth}{
\centering
\textit{Use one word from the set \{dictionary\} to describe the following piece of code: \{code\}. Also, output your confidence in your choice as a probability between 0 and 1. Don’t provide any other description. The reply form should be "word (confidence)."}}}

This prompting strategy turns out to be unsuccessful. In practice, the confidences predicted by the model were not trustworthy.
To evaluate the reliability of the model's confidence metrics, we executed the same prompt repeatedly using an exceptionally low temperature. 
In the context of LLMs, the temperature is a scaling factor applied to the token logits before converting them to token probabilities. An LLM produces the next output token by sampling from this distribution of next tokens. 
A lower temperature value makes the model's outputs more deterministic, emphasizing the most probable outcomes, whereas a higher value increases randomness, resulting in diverse outputs. 
By employing a low temperature, our objective was to curb the model's inherent randomness, leading to more deterministic outputs.

We initially hypothesized that, when the model is repeatedly invoked on the same input, if the predicted confidences are well-calibrated, outputs with higher predicted confidence values would manifest as consistent model responses while those with lower confidence would display higher variability.
Contrary to our expectations, the model displayed surprising behavior. 
For a given fixed code snippet A, the model consistently produced the same function name across multiple runs, regardless of fluctuating predicted confidence levels.
However, for another snippet B, the model's outputs varied significantly across trials, even though the expressed confidence levels remained comparably high.
Such incongruities led us to infer that the model’s confidence metrics were not dependable indicators of the likelihood of the output being correct and therefore, this prompting strategy was not viable.

\subsection{Adding Abstain Option to Model Output}
\label{abstention}
\begin{figure*}[t]
    \centering
    \includegraphics[width=0.9\textwidth]{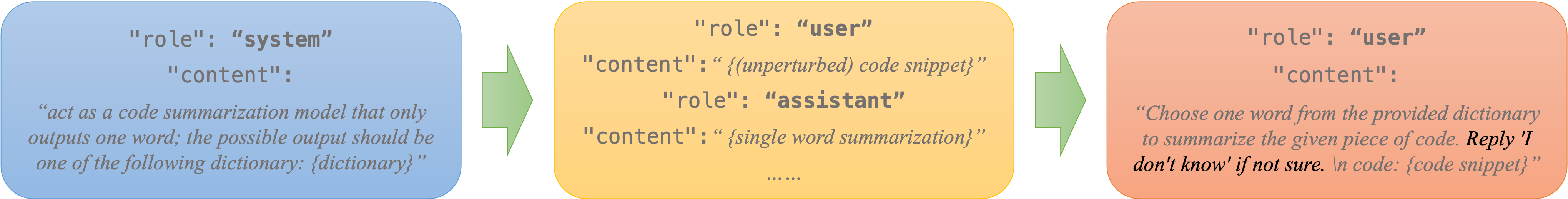}
    \caption{LLM prompt to allow the model to abstain. The \textbf{bold} text in the rightmost box shows the text that was added to the prompt from Figure~\ref{fig:summary} to allow the model to abstain from making predictions.}
    \label{fig:abs}
\end{figure*}

Although the model fails to predict its confidence score accurately, it might still be the case that the model is able to distinguish between situations where it is confident and where it is very uncertain. 
If this were true, it could suggest that in situations where the model is uncertain, the model might essentially be resorting to an arbitrary choice since it is compelled to provide an outcome.
To give the model an escape hatch in such cases of high uncertainty, 
we experimented with updating the prompt so that the model is allowed to abstain from predictions during moments of indecision, as shown in Figure~\ref{fig:abs}. 
To this end, we empower the model to output \textit{``I don't know''} whenever it encounters scenarios evoking doubt in its predictions. When exposed to adversarial examples, ideally, a model should respond \textit{``I don't know''} rather than mis-classify them. This strategy turns out to be only partially successful, as we discuss in Section~\ref{sec:experiments}.

\section{Transfer Attacks and Defenses}
\label{sec:Transfer-Defend}

\subsection{Transferability of Adversarial Examples}

The goal of this work is to assess the robustness of LLMs in performing the code summarization task when faced with malicious inputs designed to induce misclassification. 
To this end, we investigate the transferability of adversarial examples as crafted by a state-of-the-art white-box attack on the more specialized, smaller code model seq2seq~\cite{srikant2021generating, alon2018codeseq}. 
The attack is described in Section~\ref{sec:background}.

For this assessment, we consider LLMs as classifiers assigned to the task of code summarization, as described in Section~\ref{sec:LLM-code-summary} and we expose them to the adversarial examples generated from the smaller model. 
As we will demonstrate in Section~\ref{sec:experiments}, our findings indicate that a significant proportion of adversarial examples, initially generated based on seq2seq models, are transferable to LLMs, thereby highlighting potential vulnerabilities in those large language models. In the rest of this section we discuss defenses against these transfer attacks.

\subsection{Prompt-based Defenses for LLMs}\label{sec:manual-defense}
Defenses against adversarial examples typically involve re-training the model~\cite{madry2019deep} in an adversary-aware, which would be prohibitively expensive for LLMs.  In fact, due to the black-box nature of proprietary LLMs, we do not have even access to the model weights to be able to re-train the model. 
We instead propose cost-effective defenses that are prompt-based, i.e., defenses that modify the prompt to defend against adversarial attacks. The motivation for such defenses is that LLMs have shown remarkable capability to understand natural language instructions and to perform in-context learning~\cite{brown2020language}. These capabilities, combined with their ability to understand code, can be leveraged to design prompts that instruct the model to reverse the semantics-preserving code perturbations that an adversary might apply before summarizing the code. For our prompt-based defenses, we assume that the capabilities of the adversary (i.e., the kinds of semantics-preserving perturbations they can apply) are known, and also that we have access to some adversarially perturbed code samples along with their correct function names. These are standard assumptions made by the machine learning community when designing defenses against adversarial examples.

We experiment with two different approaches: defense via few-shot examples (\fsd) and defense via inverse transformation (\invd). In both cases, we design a prompt with defensive capabilities. Later, in Section~\ref{sec:metaprompt}, we present an approach that we refer to as \emph{meta-prompting} where we ask the LLM itself to generate a prompt that can help defend adversarial examples. Quite remarkably, it turns out that the LLM-generated defensive prompt can outperform the manually-generated prompts. However, in the rest of this section, we describe the two types of defensive prompts that we manually designed to improve the model's robustness to adversarial code perturbations.

\subsubsection{Defense via Few-Shot Examples (\fsd)}
\label{fsd}

\begin{figure*}[t]
    \centering
    \includegraphics[width=0.9\textwidth]{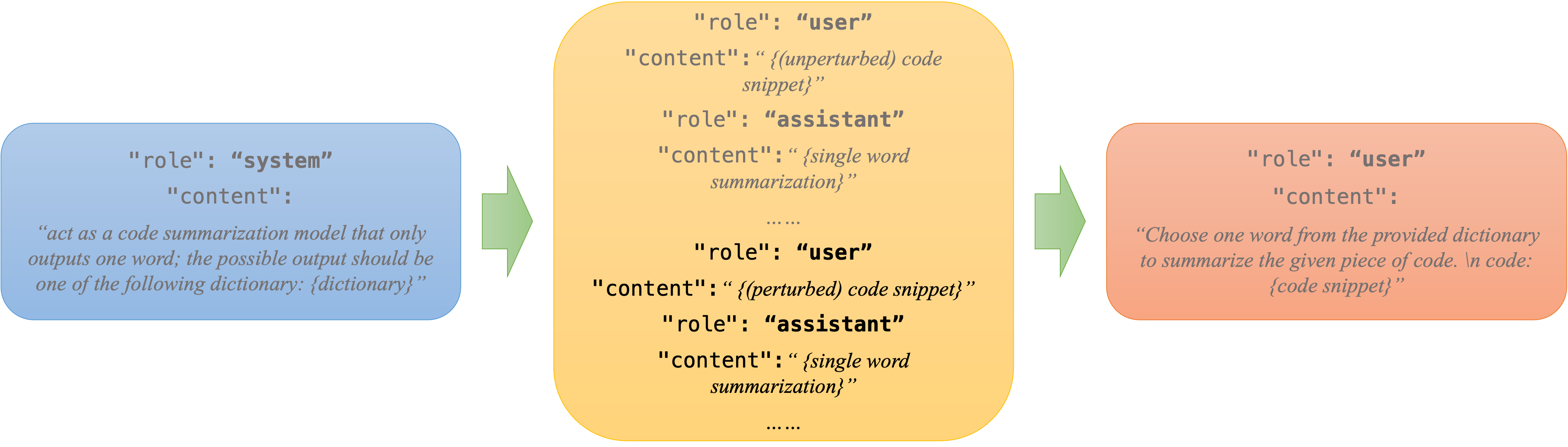}
    \caption{Prompt-based defense via few-shot examples (\fsd). The \textbf{bold} text in the middle box shows the text that was added to the prompt from Figure~\ref{fig:summary} to help the model defend against adversarial attacks.}
    \label{fig:fsd}
\end{figure*}

Building upon the strategies discussed in Section~\ref{subsec:few-shot}, where we incorporate few-shot learning in LLMs for the code summarization task, in this subsection, we explore a similar approach as a defense mechanism. 
Specifically, in addition to few-shot examples that represent unperturbed code snippets and the corresponding function names, we also incorporate typical adversarial examples (i.e., perturbed code snippets) with the correct function names as few-shot examples in the model prompt (shown in Figure~\ref{fig:fsd}). 

This approach is grounded in the hypothesis that exposing the model to adversarial examples during the few-shot learning phase can enhance its resilience against such inputs, allowing it to better generalize and correctly classify perturbed data. 
By providing the model with a more diverse set of examples, including both genuine and adversarial inputs, we aim to equip the LLMs with the necessary knowledge to recognize and counteract adversarial manipulations, thereby improving their robustness and reliability in the face of malicious attacks.

\subsubsection{Defense via Inverse Transformation (\invd)} 
\label{invd}

\begin{figure*}[t]
    \centering
    \includegraphics[width=0.9\textwidth]{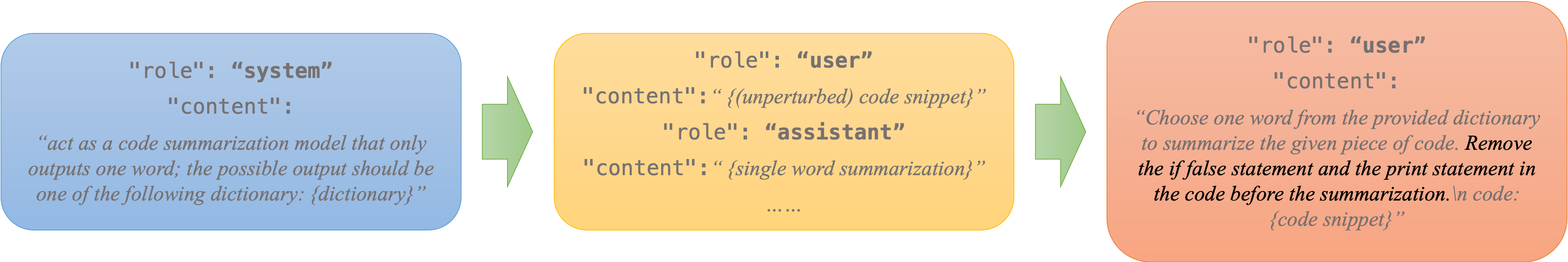}
    \caption{Prompt-based defense via inverse transformation (\invd). The \textbf{bold} text in the rightmost box shows the text that was added to the prompt from Figure~\ref{fig:summary} to help the model defend against adversarial attacks.}
    \label{fig:invd}
\end{figure*}

In this subsection, we introduce a novel self-defense technique via prompting that is both cost-effective and broadly applicable. 
This strategy utilizes the inherent capabilities of LLMs to understand instructions.
The method involves giving prompt instructions to the LLM to guide its behavior in a way that neutralizes the effects of the adversarial attack. Specifically, given the nature of code perturbations that the adversary can perform, the prompt includes instructions for undoing or inverting such perturbations.
For example, prompts can be crafted to instruct LLMs to disregard or eliminate {\em dead code} and unnecessary print statements. 
This can also extend to standardizing variable names to a canonical form as a countermeasure against variable renaming attacks. Figure~\ref{fig:invd} shows the prompt we designed for this purpose.

\paragraph{Two-step Defense.} Inspired by the effectiveness of Chain-of-Thought (CoT) prompting~\cite{wei2022chain}, we initially explored a two-step approach for defense via inverse transformation. CoT prompting is a method for improving the capability of LLMs to perform complex reasoning tasks by breaking them down into simpler sub-problems. The method involves asking the LLM to generate intermediate steps of reasoning and feeding back these LLM-generated intermediate steps into the prompt for additional context. 
In our case, we broke down the code summarization task into two steps.
The first step entailed asking the model to invert the perturbed code to its original, unaltered state. 
In the second step, the model was asked to summarize this reconstituted code to produce the final classification result. 
Each step was guided by specialized prompts designed to direct the LLM in accomplishing these tasks. 
During the first step of unperturbing the code, we carefully evaluated the capability of our prompts in guiding the LLM to accurately restore the original code from the adversarially perturbed version. 
If the prompt was found to be ineffective, it was fine-tuned iteratively to achieve better performance.

We observed that for instances of adversarially perturbed data, where the unperturbed version is correctly labeled by the LLM, the effect of the two-step inverse transformation can be categorized into three cases: \textbf{Full Success Case}, \textbf{Partial Success Case}, and \textbf{Failure Case}.

\begin{itemize}
    \item \textbf{Full Success Case:} In this case, the prompt defense based on inverse transformation proves effective both in correctly unperturbing the code and in ensuring that the model assigns the correct function name to the code snippet. This outcome represents the ideal scenario where the prompt-based defense robustly counters adversarial attacks.
    
    \item \textbf{Partial Success Case:} In this case, while the first step of the defense effectively undoes adversarial perturbations and retrieves the original source code, the model is still unable to correctly classify the code snippet, i.e., the two-step defense is not effective in ensuring correct classification. This points to a potential need for further optimization of the prompt used for the second step to harmoniously integrate the code transformation and classification steps.
    
    \item \textbf{Failure Case:} In this category, the prompt-based defense struggles to even revert the perturbed code to the original unperturbed version, and, as a consequence, the code snippet is not labeled accurately by the classifier.
\end{itemize}

The first scenario represents the ideal case, demonstrating that prompt-based defenses can indeed safeguard against adversarial examples. The second scenario highlights the need for refining our approach to identify a more synergistic prompt that can merge the two steps effectively. However, the third step also demonstrates the perils of this approach---the LLM cannot always successfully transform the code back to the original version even if the nature of code perturbations that the adversary can potentially perform is included as a part of the prompt.

Listings~\ref{code:gpt-3.5-manual} and \ref{code:gpt-4-manual} show the responses of GPT-3.5 and GPT-4, respectively, after the first step of this two-step defense. We see that GPT-4 is able to remove all the perturbations added by the adversary whereas GPT-3.5 fails to remove one of the code modifications (highlighted in gray). To invert the perturbed code, the models were given the following prompt with \textit{\{code snippet\}} replaced by the perturbed code in Listing~\ref{code:Adv}:

\noindent\fbox{%
    \parbox{\textwidth}{
    \centering
    \textit{Remove the if false statement and the print statement in the code before the summarization.\textbackslash n code: \{code snippet\}}}}

\paragraph{Single-step Defense.} Following the initial exploration with the two-step defense, we later refined our strategy by consolidating it into a single comprehensive prompt. A single-step approach is more cost-effective since it requires fewer queries to the LLM.
The unified prompt directs the LLM to, both, invert the adversarial code transformations with the hope of recovering the original code and producing the final classification result in a single pass. In other words, LLM is now guided first to recover the original code and then directly output only the final classification result, all under the guidance of this integrated prompt. The model never explicitly outputs the inverted code. Figure~\ref{fig:invd} shows the prompt we used for this single-step approach. 
As stated earlier, this consolidation aims to improve cost effectiveness while maintaining the robustness of our defensive strategy. In Section~\ref{sec:experiments}, we present results demonstrating the effectiveness of the single-step inverse transformation defense.

\begin{lstlisting}[language=Python, caption={GPT-3.5 generated response based on crafted prompt for \invd~defense.}, label=code:gpt-3.5-manual, float=t]
code: ( self application name = python gntp notifications = [ ] default notifications = none application icon = none hostname = localhost password = none port = 23053 ) :    self . application name = application name    self . notifications = list ( notifications )    if default notifications : @\mytikzmark{hl5Start}@traverse = 1@\mytikzmark{hl5End}@   self . default notifications = list ( default notifications )    else : self . default notifications = self . notifications    self . application icon = application icon    self . password = password    self . hostname = hostname    self . port = int ( port )
 \end{lstlisting}

 \begin{tikzpicture}[remember picture, overlay]
        \highlight{hl5Start}{hl5End}
    \end{tikzpicture} 

 \begin{lstlisting}[language=Python, caption={GPT-4 generated response based on crafted prompt for \invd~defense.}, label=code:gpt-4-manual, float=t]
code: ( self application name = python gntp notifications = [ ] default notifications = none application icon = none hostname = localhost password = none port = 23053 ) :    self . application name = application name    self . notifications = list ( notifications )    if default notifications : self . default notifications = list ( default notifications )    else : self . default notifications = self . notifications    self . application icon = application icon    self . password = password    self . hostname = hostname    self . port = int ( port )
\end{lstlisting}

\section{Meta-Prompting: LLM-Generated Defensive Prompts}
\label{sec:metaprompt}

The effectiveness of our prompt-based defenses hinges on the the quality of the prompts used to instruct the LLMs to neutralize known adversarial attacks. Leveraging the generative capabilities of LLMs we experimented with a technique that we call {\em meta-prompting}, i.e., we instructed an LLM, namely GPT-4,  to generate the prompt defense by itself. As it turns out, the LLM generated prompts seem much more effective than the manually engineered prompts in defending against adversarial attacks.

\begin{figure*}[t]
    \centering
    \includegraphics[width=0.7\textwidth]{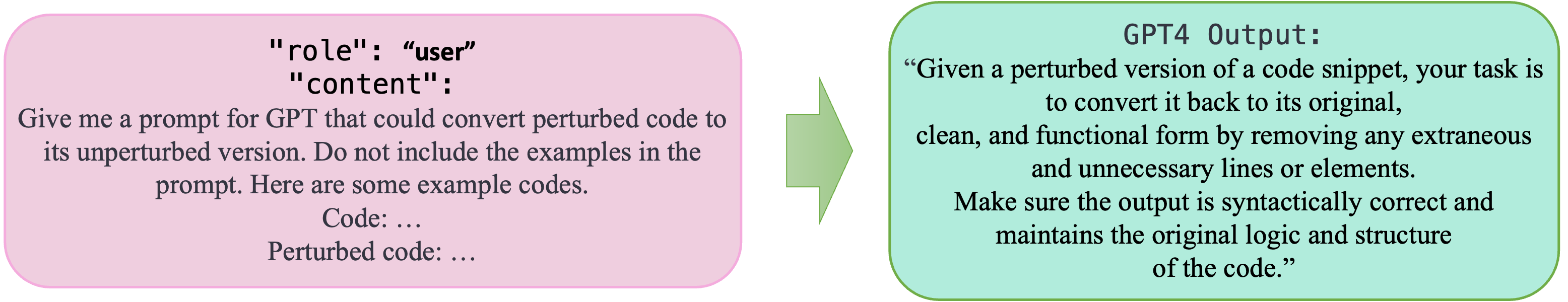}
\caption{Example-based meta-prompting}
    \label{fig:EBMP}
\end{figure*}

\begin{figure*}[t]
    \centering
    \includegraphics[width=0.9\textwidth]{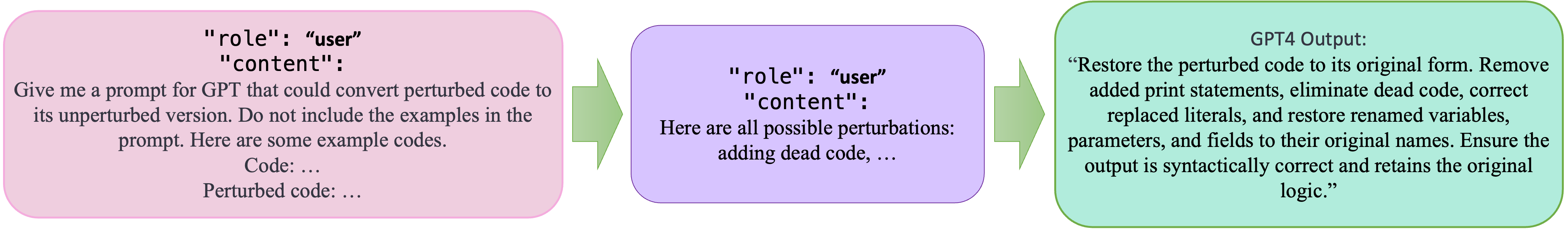}%
    \caption{Perturbation-aware meta-prompting}
    \label{fig:PAMP}
\end{figure*}

Meta-prompting is achieved by feeding the model higher-level instructions or templates, along with examples of both original (unperturbed) and perturbed code. 
The LLM, utilizing these inputs, generates effective prompts tailored to the specific nature of the perturbations and the desired outcome. 
We experimented with two techniques, as illustrated in Figures~\ref{fig:EBMP} and~\ref{fig:PAMP}.
\begin{enumerate}
    \item \textbf{Example-Based Meta-Prompting (EBMP):} 
    This technique involves providing the LLM with a couple of examples of original and corresponding perturbed code, without any explicit insight into the nature of the perturbations. 
    The LLM is instructed to analyze these examples and generate prompts that could effectively guide the inverse transformation process. 
    This method relies on the LLM's own capabilities to deduce the necessary transformations.
    The generated prompt is as follows: 
    
    \noindent\fbox{%
    \parbox{\dimexpr\textwidth-\leftmargin-\itemindent-2\fboxsep}{
    \centering
    \textit{"Given a perturbed version of a code snippet, your task is to convert it back to its original, clean, and functional form by removing any extraneous and unnecessary lines or elements. Make sure the output is syntactically correct and maintains the original logic and structure of the code."}}}

    \item \textbf{Perturbation-Aware Meta-Prompting (PAMP):} 
    The second technique assumes some knowledge about the possible perturbations. 
    This allows the LLM to create a specialized prompt that is more aligned with the specific code transformations employed by an adversary. 
    This method is anticipated to be more effective but requires prior knowledge or assumptions about the attacker's perturbations.
    The generated prompt is as follows: 
    \noindent\fbox{%
    \parbox{\dimexpr\textwidth-\leftmargin-\itemindent-2\fboxsep}{
    \centering
    \textit{"Restore the perturbed code to its original form. Remove added print statements, eliminate dead code, correct replaced literals, and restore renamed variables, parameters, and fields to their original names. Ensure the output is syntactically correct and retains the original logic."}}}
    
\end{enumerate}

These LLM-generated prompts are incorporated into our prompt-based defense via inverse transformation (Section~\ref{invd} and Figure~\ref{fig:invd}). 
In particular, the following instruction in Figure~\ref{fig:invd}

\noindent\fbox{\parbox{\textwidth}{
\centering
    \textit{"Remove the if false statement and the print statement in the code before the summarization."}}}

\noindent is replaced by the instruction 
\fbox{
    \textit{"Before summarization, \{LLM-generated prompt\}."}} 
where \textit{\{LLM-generated prompt\}} is substituted by the prompt generated via meta-prompting.
The other parts of the prompt from Figure~\ref{fig:invd} remain unchanged.

Note the LLM-generated prompts include additional instructions that were not included in the manually crafted prompt either by design or due to inadvertent omission. In particular, the last sentences in both the prompts instructing that the output should be syntactically correct and retain the original logic of the code are, in hindsight, useful instructions that should have been included in the manually crafted prompt as well. We also note that the perturbation-aware nature of PAMP leads it to generate a prompt that includes even more suggestions for reverting the code to its original form than in our hand-crafted prompt. In this case, however, the omissions are by design---a manually crafted prompt that instructed the model to rename variables and parameters was less effective than our final defensive prompt presented in Figure~\ref{fig:invd}.

\subsection{Example Comparing GPT-3.5 vs. GPT-4 using the Generated Prompts}
\label{sec:compare_meta}

\begin{table*}[t]
\noindent\begin{minipage}{.45\textwidth}
\begin{lstlisting}[language=Python, caption={GPT-3.5 generated response via EBMP.}, label=code:GPT3.5-EBMP]
self.application_name = application_name
self.notifications = list(notifications)
if default_notifications:
    @\mytikzmark{hl6Start}@traverse = 1@\mytikzmark{hl6End}@
    self.default_notifications = list(default_notifications)
else:
    self.default_notifications = self.notifications
self.application_icon = application_icon
@\mytikzmark{hl7Start}@print(pspace)@\mytikzmark{hl7End}@
self.password = password
self.hostname = hostname
@\mytikzmark{hl8Start}@validity = 1@\mytikzmark{hl8End}@
self.port = int(port)
@\mytikzmark{hl9Start}@sl = 1@\mytikzmark{hl9End}@
@\mytikzmark{hl10Start}@evoked = 1@\mytikzmark{hl10End}@
\end{lstlisting}
\end{minipage}\hfill
\begin{minipage}{.45\textwidth}
\begin{lstlisting}[language=Python, caption={GPT-3.5 generated response via PAMP.}, label=code:GPT3.5-PAMP]
class Application:
    def __init__(self, application_name, notifications, default_notifications, application_icon, hostname, password, port):
        self.application_name = application_name
        self.notifications = list(notificatiions)
        if default_notifications:
            self.default_notifications = list(default_notifications)
        else:
            self.default_notifications = self.notifications
        self.application_icon = application_icon
        self.password = password
        self.hostname = hostname
        self.port = int(port)
\end{lstlisting}
\end{minipage}

 \begin{tikzpicture}[remember picture, overlay]
        \highlight{hl6Start}{hl6End}
        \highlight{hl7Start}{hl7End}
        \highlight{hl8Start}{hl8End}
        \highlight{hl9Start}{hl9End}
        \highlight{hl10Start}{hl10End}
    \end{tikzpicture} 

\noindent\begin{minipage}{.45\textwidth}
\begin{lstlisting}[language=Python, caption=GPT-4 generated response via EBMP., label=code:GPT4-EBMP]
The cleaned code should look like this:

```python
class MyClass:
    def __init__(self, application_name, notifications=[], default_notifications=None, application_icon=None, hostname='localhost', password=None, port=23053):
        self.application_name = application_name
        self.notifications = list(notifications)
        if default_notifications:
            self.default_notifications = list(default_notifications)
        else:
            self.default_notifications = self.notifications
        self.application_icon = application_icon
        self.password = password
        self.hostname = hostname
        self.port = int(port)
```

I assumed that this code is a part of a class, because it seems to be a constructor method (it initializes instance variables). I also removed all the unnecessary lines and elements.
\end{lstlisting}
\end{minipage}\hfill
\begin{minipage}{.45\textwidth}
\begin{lstlisting}[language=Python, caption=GPT-4 generated response via PAMP., label=code:GPT4-PAMP]
The restored code should look like this:

```python
def __init__(self, application_name='python gntp', notifications=[], default_notifications=None, application_icon=None, hostname='localhost', password=None, port=23053):
    self.application_name = application_name
    self.notifications = list(notifications)
    if default_notifications:
        self.default_notifications = list(default_notifications)
    else:
        self.default_notifications = self.notifications
    self.application_icon = application_icon
    self.password = password
    self.hostname = hostname
    self.port = int(port)
```

In the restored code, I have removed all the unnecessary print statements and dead code (code that can never be executed). I have also corrected the function definition line to make it a valid Python function. The parameters have been renamed to their original names and the function now has a name (`__init__`). The literals have been corrected to their original values.
\end{lstlisting}
\end{minipage}
\end{table*}
We discuss here the effect on issuing the meta-generated prompts to the GPT-3.5 and GPT-4 models on the perturbed example shown in Listings \ref{code:Adv} that was generated by adverarially perturbing a randomly chosen sample (shown in Listing~\ref{code:original}) from our dataset. 
The effect of applying the prompts generated via Example-Based Meta-Prompting and the Perturbation-Aware Meta-Prompting to the adversarial example is illustrated in Listings~\ref{code:GPT3.5-EBMP},~\ref{code:GPT3.5-PAMP} (for GPT-3.5) and Listings~\ref{code:GPT4-EBMP},~\ref{code:GPT4-PAMP} (for GPT-4). Note that these results are generated by directly issuing the prompts shown in the rightmost boxes of Figures~\ref{fig:EBMP} and \ref{fig:PAMP}.

We make a number of observations. In all the cases here, GPT-3.5 and GPT-4 are able to infer the correct indentation for the code snippet even though the meta-generated prompts do not include any such instruction. Moreover, except for the GPT-3.5 and EBMP combination, the models are even able to infer a correct name for the function (\texttt{\_\_init\_\_}). In some cases (Listings~\ref{code:GPT3.5-PAMP} and \ref{code:GPT4-EBMP}), the model also invents class names. We see that the combination of GPT-3.5 and EBMP fails to remove some of the adversarial perturbations in the code. GPT-3.5 also fails to pass correct default values to the function parameters whereas GPT-4 is better at this, particularly when using the prompt generated via PAMP.  GPT-4 also tends to produce explanations for its actions. Overall, this simple exercise suggests that the combination of GPT-4 and the PAMP generated prompt may lead to the best results, as one would expect, and this is indeed what we observe in our experiments. However, we note that although all these responses are generated with the temperature parameter set to 0, the generation process still exhibits some randomness, and also the behavior of the models on a different example might be different.

\section{Experiments}
\label{sec:experiments}
We report on our experiments that aim to answer the following research questions:

\begin{itemize}
\item{RQ1}: Do code attacks transfer from a small model to LLMs?
\item{RQ2}: Do the manually-engineered prompt defenses succeed against attacks?
\item{RQ3}: Does meta-prompting help? 
\item{RQ4}: How does the abstain option influence model accuracy, transfer of attacks, and defense?
\end{itemize}

\subsection{Dataset}
\label{subsec:dataset}

In Section~\ref{attack_seq2seq}, we described the methodology  for generating the adversarial dataset according to~\citet{srikant2021generating}. 
We started with a dataset for the code summarization task such that given a function body as input, the output should be a single-word label representing the function name. 
The programming language used in this dataset is Python. 
We generated adversarial examples with respect to a small code model (based on the seq2seq~\cite{sutskever2014sequence} architecture) trained on the code summarization task by applying semantics-preserving perturbations to the input code snippets as proposed in the white-box attack of \citet{srikant2021generating}. 
As highlighted in Section~\ref{attack_seq2seq}, the attack algorithm solves a joint optimization problem to address the site selection and site perturbation problems.
We used the Randomized Smoothing~\cite{duchi2012randomized} option to ease the optimization problem and improve the generation of adversarial programs, as suggested by~\citet{srikant2021generating}. 
One parameter that needs to be set for the adversarial generation process is the perturbation strength, denoted as \( k \). 
In our experiments, we set \( k = 5 \), implying that an attacker can perturb a maximum of five sites within a given program. 
This constraint ensures that the adversarial programs remain close to the original while still maintaining their adversarial intent. 
Though the original attack work by~\citet{srikant2021generating} reported results on 2800 samples (out of a Python dataset of 150K programs~\cite{raychev2016probabilistic}) whose unperturbed versions are correctly labeled by the small model (seq2seq),%
we randomly sampled 1000 examples from this set for our experiments; this is due to the significant 
computational resources required to run experiments with state-of-the-art LLMs. We use $S$ to denote this dataset of 1000 unperturbed samples.

\subsection{Models}

As mentioned, we use a small code model (with 48M parameters)
trained on the code summarization task that has a seq2seq~\cite{sutskever2014sequence} architecture to generate the attacks. 
This \textbf{seq2seq} model consists of two main components: an encoder that processes the input sequence and compresses it into a fixed-length vector (context), and a decoder that generates an output sequence based on this context.
We evaluate the attack generated from the seq2seq model on the proprietary and open-source LLMs that are described below. We set temperature to 0 for all our experiments to control the randomness of the models.

\paragraph{Proprietary Models.} We evaluate state-of-the-art proprietary LLMs, including \textbf{GPT-3.5} (\texttt{gpt-3.5-\linebreak turbo-0613}) and \textbf{GPT-4} (\texttt{gpt-4-0613}) from OpenAI~\cite{openai2023gpt4}, and \textbf{Claude-Instant-1} (\texttt{claude-instant-1}) and \textbf{Claude-2} from Anthropic~\cite{anthropic}, on the generated attacks. These models are amongst the top-5 entries on a public leaderboard\footnote{\url{https://chat.lmsys.org} (Accessed in November, 2023)} that tracks the performance of LLMs on multiple tasks. GPT-3.5 and GPT-4 have been shown to possess remarkable abilities on various human tasks, including code understanding and code co-piloting. On the other hand, Claude models are able to process long sequences (up to 100k tokens) and have been shown as the most robust models under a state-of-the-art attack~\cite{UniversalAttack23} that aims to elicit harmful knowledge from the models. In our experiments,  we focus on their ability to perform code understanding and assess their response in the presence of adversarial code examples.

\paragraph{Open-Source Models.}  We also evaluate \textbf{CodeLlama}~\cite{roziere2023code}, a state-of-the-art open-source LLM trained to follow instructions related to coding problems. CodeLlama is fine-tuned from the weights of Llama-2~\cite{touvron2023llama}, an LLM released by Meta in early July 2023. While Llama-2 is developed to follow general instructions, CodeLlama is further fine-tuned on code datasets so is more capable of understanding code compared to the original Llama-2. There are three different sizes of CodeLlama (i.e. with 7B, 13B and 34B parameters) and two tuning options (i.e. with or without instruction-following tuning to understand natural language inputs). In our experiment, we use \texttt{CodeLlama-7B-Instruct}, which has 7B parameters and is tuned to answer coding questions written in natural language, e.g. summarizing the purpose of a given function.

\subsection{Notations}
We use $S$ to denote the set of 1000 clean inputs that are correctly classified by the baseline seq2seq model. We use $S_M$ to denote the subset of $S$ that each model $M$ correctly classifies. For instance, when $M=$GPT-3.5, $S_M$ denotes the subset of $S$ that GPT-3.5 classifies correctly. When $M=$seq2seq, $S_M=S$. We compute $S_M$ in this way to allow for a fair comparison among the models, as 
~\citet{srikant2021generating} apply adversarial code perturbations only to the data that is classified correctly by the model. We use $S_M^{adv}$ to denote the set of adversarial inputs obtained by applying the perturbations on inputs in $S_M$; note that the sizes of $S_M$ and $S_M^{adv}$ are the same, i.e., $|S_M|=|S_M^{adv}|$. For some dataset $X$, we use notations $Correct(X)$, $Wrong(X)$, $Abstain(X)$ to denote the number of instances in $X$ for which the output of a model is correctly classified, mis-classified, or {\em I don't know}, respectively, where the model is implicit from the context.

\subsection{RQ1: Do code attacks transfer from a small model to LLMs?}

\begin{table}[t]
\centering\caption{Transferability of the attack across different models.} 
\label{tab:comparison_asr}
\begin{tabular}{|l|c|c|}
\hline
\textbf{Model ($M$)} & \textbf{Acc on $S$} & \textbf{ASR} \\
 & \scriptsize\textbf{($Correct(S) / |S|$)}  & \scriptsize\textbf{($Wrong(S_M^{adv}) / |S_M|$)} \\
\hline
\textbf{seq2seq}  & 100 & 44.76 \\
\hline
\textbf{GPT-3.5} & 60.7 & 29.49\\
\hline
\textbf{GPT-4} & 67 & 21.04\\
\hline
\textbf{Claude-Instant-1} & 47.4 & 25.11 \\
\hline
\textbf{Claude-2} & 59.6 & 22.99 \\
\hline
\textbf{CodeLlama} & 9 & 53.33\\
\hline
\end{tabular}%
\end{table}

To investigate the transferability of adversarial attacks from our baseline model, seq2seq, we ran the various LLMs on the adversarial dataset, as well as on the corresponding clean, unperturbed dataset, for comparison; the results are reported in Table~\ref{tab:comparison_asr}.
The table contrasts the accuracy of each model on the clean dataset, \( S \), with the Attack Success Rate (ASR)~\cite{srikant2021generating}, which are defined below. 

\begin{itemize}
\item Given a model $M$ and a dataset $S$, the accuracy {\bf Acc on} $S$ is defined as $Correct(S)/|S|$.
\item Given a model $M$ and dataset $S_M$ of clean inputs that are correctly classfied by $M$, the {\bf Attack Success Rate (ASR)} is defined as $Wrong(S_M^{adv})/|S_M|$.
\end{itemize}

\( S \), as discussed, is a Python dataset with 1000 clean inputs (sampled from a dataset with a 150K programs) that are correctly classified by the seq2seq model and on which the attack was applied. Therefore, the accuracy of the seq2seq model on $S$ is 100\%.
However, the overall accuracy of this seq2seq model on the entire dataset of Python programs is quite low, around 25\%.

From our data, the seq2seq model has an ASR of 44.76\%. The attack was able to deceive the seq2seq model for nearly half of the adversarial samples effectively.
The results indicate that the attacks from the baseline model, seq2seq, exhibit varying degrees of transferability across different LLMs. 
For instance, GPT-3.5, which has an accuracy of 60.7\% on $S$, presents an ASR of 29.49\%, suggesting a notably large transferability on this attack. 
GPT-4 has the highest accuracy of 67\% and the lowest ASR at 21.04\%, which is still significant.
The Claude models, Claude-Instant-1 and Claude-2, with the accuracy of 47.4\% and 59.6\%, respectively, display ASRs of 25.11\% and 22.99\%, indicating that they are also quite vulnerable to attacks.  

Different from other LLMs, CodeLlama displays the lowest accuracy of 9\% and shows the highest ASR of 53.33\%, indicating a pronounced susceptibility to attacks from the seq2seq model. 
Generally, CodeLlama underperforms compared to other models. We believe this is primarily due to its tendency to generate lengthy explanations of the input code snippet rather than succinctly summarizing it with a single keyword. 
Although we assess CodeLlama's output based on the inclusion of target words, its performance still falls short of other LLMs in terms of output quality and precision. Therefore, the results that we report on this model throughout the paper need to be taken with a grain of salt, as the model was not well-suited for the code summarization task. 

In summary, the attacks transfer from smaller models such as seq2seq to all the larger models with significant attack success rates.
These results indicate that one can effectively attack these black-box, very large, performant LLMs by simply generating adversarial examples based on much smaller models. This makes it significantly more feasible for attackers to succeed.

\subsection{RQ2: Do the manually-engineered prompt defenses succeed against attacks?}
\label{sec:rq2}
\begin{table*}[t]
\centering
\caption{Effectiveness of the defenses across different models.}
\label{tab:model_comparison_defense}
\resizebox{\textwidth}{!}{%
\begin{tabular}{|l|c|c|c|c|}
\hline
\textbf{Model ($M$)}& \textbf{\fsd~ Acc on $S_M$} & \textbf{\fsd~ASR} & \textbf{\invd~Acc on $S_M$} & \textbf{\invd~ASR} \\
  & \scriptsize\textbf{$(Correct(S_M) / |S_M|$)} & \scriptsize\textbf{$(Wrong(S_M^{adv}) / |S_M|$)} &  \scriptsize\textbf{$(Correct(S_M) / |S_M|$)} & \scriptsize\textbf{$(Wrong(S_M^{adv}) / |S_M|$)} \\ 
\hline
\textbf{GPT-3.5} & 93.57 & 30.15 & 92.59 & 24.71 \\ %
\hline
\textbf{GPT-4} & 91.94 & 14.63 & 91.79 & 16.57\\
\hline
\textbf{Claude-Instant-1} & 90.51 & 9.70 & 79.75 & 19.20 \\
\hline
\textbf{Claude-2} & 86.58 & 10.91 & 82.38 & 20.97\\
\hline
\textbf{CodeLlama} & 74.44 & 66.67 & 50 & 74.44 \\
\hline
\end{tabular}%
}
\end{table*}

To evaluate the proposed defenses, \(\fsd\) (Defense via Few-Shot Examples, Section~\ref{fsd}) and \(\invd\) (Defense via Inverse Transformation, Section~\ref{invd}), we first compute the accuracy of the model with the defense on clean, unperturbed inputs. Specifically, we measure the percentage of clean inputs (out of $S_M$ for each model $M$) that remain correctly labeled after the application of the defense (denoted as accuracies $\fsd$ {\bf Acc on $S_M$} and $\invd$ {\bf Acc on $S_M$} in the table). This is to determine if the defense does not unintentionally fool the model into misclassifying clean inputs that would otherwise be correctly classified by the model without any defense.

To answer the main question of defense effectiveness against attacks, we also compare the Attack Success Rate (ASR) for the models without defenses against their ASR when these defenses are applied.
The results are presented in Table \ref{tab:model_comparison_defense}.

We first note that the accuracy (Acc on $S_M$) remains high when the defenses are applied, indicating that defenses do not accidentally lead to too many misclassifications of clean inputs; the outlier is again CodeLlama.

In terms of defense effectiveness, we note that for the GPT-3.5 model, the ASR reduces from an original 29.49\% (computed on the model without defense; see Table~\ref{tab:comparison_asr}) to 24.71\% using the \(\invd\) defense, indicating its effectiveness. 
However, the \(\fsd\) defense seems not to be effective, as the ASR is approximately unchanged (in fact, slightly larger). 
In the context of the GPT-4 model, both defenses curtail the ASR from the baseline 21.04\% (model without defense; see Table~\ref{tab:comparison_asr}), with \(\fsd\) exhibiting superior performance (ASR 14.63\%). 
The Claude-Instant-1 and Claude-2 models both benefit from the defenses, with ASRs decreasing from their initial values of 25.11\% and 22.99\%, respectively on the models without defense, to $\fsd$ ASR of 9.7\% and 10.91\%; again, the \(\fsd\) defense is more effective. 
For the CodeLlama model, while the ASR diminishes from an initial 53.33\% with the \(\fsd\) defense, the \(\invd\) defense markedly amplifies it.

In summary, the \(\fsd\) defense appears to generally improve the models' resilience against adversarial attacks. 
However, the effectiveness of the \(\invd\) defense is less clear and appears to be model-dependent.

\subsection{RQ3: Does meta-prompting help?}

\begin{table}[t]
\centering
\caption{Evaluating \invd~defense with LLM-generated prompts}
\begin{tabular}{|l|c|c|c|c|c|}
\hline
 \multirow{2}{*}{\textbf{Model($M$)}}&  \multicolumn{2}{|c|}{\textbf{EBMP~+~\invd}}  &  \multicolumn{2}{|c|}{\textbf{PAMP~+~\invd}} \\ 
 \cline{2-5}
 & \textbf{Acc on $S_M$} & \textbf{ASR} & \textbf{Acc on $S_M$} & \textbf{ASR} \\ 
 & \scriptsize\textbf{$(Correct(S_M)/|S_M|$)} & \scriptsize \textbf{$(Wrong(S^{adv}_M)/|S_M|$)} & \scriptsize \textbf{$(Correct(S_M)/|S_M|$)} &\scriptsize \textbf{$(Wrong(S^{adv}_M)/|S_M|$)} \\ \hline
 
\textbf{GPT-3.5}  & 89.71 & 10.48 & 94.3 & 6.07 \\ \hline
\textbf{GPT-4}  & 95.4 & 4.3 & 94.96 & 3.86 \\ \hline
\end{tabular}
\label{Meta-Prompting}
\end{table}

We evaluate the integration of the two novel prompts, generated via Example-Based Meta-Prompting (EBMP) and Perturbation-Aware Meta-Prompting (PAMP), with the \invd~ method. Table~\ref{Meta-Prompting} shows our results when using GPT-3.5 and GPT-4. We report the same metrics as in Section~\ref{sec:rq2} where we evaluated the effectiveness of the manually constructed prompt-based defenses.

 We see that the effect of these defenses on the clean code samples (reported in columns \textbf{Acc on $S_M$}) is similar to the results observed for the $\invd$ defense with manually generated prompts. For instance, GPT-3.5 with a PAMP-generated prompt has an accuracy of 94.3\% on $S_M$ compared to 92.59\% with the manually generated prompt. Similarly,  GPT-4 with PAMP-generated prompt has an accuracy of 94.96\% on $S_M$ compared to 91.79\% with the manually generated prompt. %
These results suggest that EBMP and PAMP generated prompts maintain high accuracy in labeling non-adversarial examples while implementing defensive measures.

More interestingly, there is a significant reduction in the Attack Success Rate (ASR) when employing these techniques, indicating that the LLM generated prompts can provide a strong defense against adversarial attacks. 
Notably, using the EBMP generated prompt, all of the samples whose clean code versions are correctly labeled continue to be correctly labeled for their corresponding adversarially perturbed code versions, i.e., the attack fails to have any effect. %
This implies a near-elimination of the effects of adversarial perturbations which we find particularly striking.

PAMP-generated prompts lead to the lowest ASR---6.07\% for GPT-3.5 and 3.86\% for GPT-4, compared to 24.71\% and 16.57\%, respectively, when using the manually-crafted prompt.
This improved efficacy can be attributed to PAMP's design, 
which incorporates information about potential perturbations an attacker might employ. The observation that the combination of PAMP and GPT-4 leads to the best results is in sync with the effects on the example code snippet that we discussed in Section~\ref{sec:compare_meta}.

\paragraph{Discussion.} We find it intriguing that LLM-generated prompts are more effective than the manually crafted prompts at defending the model from adversarial perturbations. In terms of the actual content of the prompt, the LLM-generated prompts are almost identical to the manual prompts except for some additional instructions, in particular, the instruction to ensure that the output is syntactically correct and retains the original logic that appears both in the EBMP and PAMP generated prompts. We hypothesize that the effectiveness of meta-prompting is an instance of a generally observed surprising capability of LLMs to generate prompts that are more effective that ones crafted by humans~\cite{zhou2023large,pryzant2023automatic}. 

Note that due to resource constraints, we were only able to evaluate the GPT-3.5 and GPT-4 models. Further evaluation with the other LLMs is left for future work.

\subsection{RQ4: How does the abstain option influence model accuracy, transfer of attacks, and defense?}
\label{subsec:RQ2}

\begin{table*}[t]
\centering
\caption{Effect of abstain option on transfer of attack.}
\label{tab:idk_adv}
\begin{tabular}{|l|c|c|c|c|}
\hline
\multirow{2}{*}{\textbf{Model ($M$)}}& \multicolumn{2}{|c|}{Experiment 1: Clean Samples} & \multicolumn{2}{|c|}{Experiment 2: Adversarial Samples} \\
\cline{2-5}
    & \textbf{Abs on $S$} & \textbf{Acc on $S$} 
    &  \textbf{Abs on $S_M^{adv}$} & \textbf{ASR} 
    \\ 
    & \scriptsize\textbf{$(Abstain(S)/|S|$)} 
    & \scriptsize\textbf{$(Correct(S)/|S|$)} %
    &  \scriptsize\textbf{$(Abstain(S_M^{adv})/|S_M|$)} 
    & \scriptsize\textbf{$(Wrong(S_M^{adv})/|S_M|$)} %
    \\ 
\hline
\textbf{GPT-3.5} & 27.8 & 40.6 %
& 83.20 & 6.43 %
\\
\hline
\textbf{GPT-4} & 0 & 64.5 %
& 0 & 22.09 %
\\
\hline
\textbf{Claude-Instant-1}  & 1 & 44.1 %
& 6.33 & 21.10 %
\\
\hline
\textbf{Claude-2}  & 2.3 & 58.1 %
& 8.05 & 19.13 %
\\
\hline
\textbf{CodeLlama}  & 59.9 & 6.5 %
& 91.11 & 1.11 %
\\
\hline
\end{tabular}%
\end{table*}

To answer this research question, we designed two experiments on the LLMs, allowing the model to abstain from giving a summarization word when the model is not sure about its answer as described in Section~\ref{abstention}.

\begin{itemize}
\item{\bf (Experiment 1:)} In the first experiment, we evaluate each LLM with the abstain option on the clean dataset \(S\) to determine how often the model abstains on clean inputs.
\item{\bf (Experiment 2:)} In the second experiment, we evaluate each LLM with the abstain option on the perturbed inputs from dataset \(S_M\). Here, \(S_M\) is the same as before, i.e., the subset of \(S\) that the LLM \(M\) (without abstain option) correctly labeled. This allows us to evaluate the effectiveness of the attack (measured by ASR) in the presence of the abstain option and compare it with the ASR computed for the models without the abstain option (Table~\ref{tab:comparison_asr}).
\end{itemize}

Table~\ref{tab:idk_adv} displays the results of these experiments. 
We use \textbf{Abs} on various datasets to denote {\bf abstain rates} which are meant to measure how often the models abstain when classifying inputs from the datasets. \textbf{Acc} is accuracy and \textbf{ASR} is attack success rate.

The results indicate that, as desired, ASR decreases for all models (when compared to ASR for the models without abstain option; see Table~\ref{tab:comparison_asr}); one exception is GPT-4. Furthermore, for GPT-3.5 and CodeLlama the abstain rate is high on adversarial inputs, indicating that the approach could be potentially useful for detecting adversarial inputs on those models. However, the drawback is that the accuracy decreases for all the models except for GPT-4, and the abstain rate may be too high on clean inputs (as in the case of GPT-3.5 and CodeLlama).

\begin{table*}[t]
\centering
\caption{Effect of abstain option for the defenses. }
\label{tab:idk_adv_fsd_pd}
\resizebox{\textwidth}{!}{%
\begin{tabular}{|l|c|c|c|c|c|c|}
\hline
\textbf{Model} 
  &\textbf{\fsd~ Abs on $S_M^{adv}$} & \textbf{\fsd~ASR} 
  &  \textbf{\invd~ Abs on $S_M^{adv}$} & \textbf{\invd~ASR} 
  \\ 
  & \scriptsize\textbf{$(Abstain(S_M^{adv})/|S_M|$)} 
  & \scriptsize\textbf{$(Wrong(S_M^{adv})/|S_M|$)} %
  &  \scriptsize\textbf{$(Abstain(S_M^{adv})/|S_M|$)} 
  & \scriptsize\textbf{$(Wrong(S_M^{adv})/|S_M|$)} %
  \\ 
\hline
\textbf{GPT-3.5} & 59.14 & 14.00 %
& 73.97 & 6.59 %
\\
\hline
\textbf{GPT-4} & 0 & 15.52 %
& 0.29 & 17.16 %
\\
\hline
\textbf{Claude-Instant-1}  & 0.42 & 11.81 %
& 0 & 23 %
\\
\hline
\textbf{Claude-2}  & 1.01 & 18.12 %
& 2.01 & 23.49 %
\\
\hline
\textbf{CodeLlama}  & 97.78 & 1.11 %
& 81.11 & 13.33 %
\\
\hline
\end{tabular}%
}
\end{table*}

The abstain mechanism offers some defense by itself (as it can flag adversarial inputs as {\em I don't know}), which is orthogonal to the other defenses ($\fsd$ and $\invd$). We also experimented with combining the abstain option with $\fsd$ and $\invd$ defenses. The results are displayed in Table~\ref{tab:idk_adv_fsd_pd}.

The experiments are inconclusive in the sense that it is not clear that the combined defenses lead to lower ASR (as compared to results in Table~\ref{tab:model_comparison_defense} obtained on defended models without abstain option). One interesting observation is that the defenses tend to reduce the confusion of the models (evidenced by lower abstain rates). This is as expected, as the two defenses $\fsd$ and $\invd$ aim to revert obfuscating transformations. 

While it is unclear that the abstain option is useful for detecting and defending against adversarial attacks, our results indicate that it can potentially be used to detect the confusion of the model. This, in turn, may be useful in other applications, such as the detection of out-of-distribution inputs, that we plan to explore.

\section{Related Work}
\label{sec:related}

Deep neural networks (DNNs) have been shown to obtain state-of-the art performance for a variety of tasks, including image classification \cite{imagenet} and natural language processing \cite{devlin2019bert}.  DNNs have also found success in solving programming language tasks, such as code summarization \cite{allamanis2016convolutional, leclair2020improved,alon2018codeseq,alon2019code2vec}, bug prediction \cite{DeepBugs,allamanis2018learning,vasic2018neural}, or program repair \cite{Hoppity,xia2023automated}. Popular pre-LLM models include code2vec~\cite{alon2019code2vec}, code2seq~\cite{alon2018codeseq}, and CodeBERT~\cite{feng2020codebert}. 

However it is known that even highly performant neural networks are vulnerable to adversarial attacks \cite{szegedy2014intriguing,GoodfellowSS14}, i.e., small perturbations to an input designed to change correct predictions of  state-of-the-art DNNs.
In natural language processing, several techniques have been proposed to generate adversarial examples in a black-box \cite{alzantot-etal-2018-generating} or white-box \cite{ebrahimi-etal-2018-hotflip} manner.
 
For code models, there are a few works that study the vulnerability of code models to adversarial examples, such as the inclusion of ``dead code'' through false conditions or the addition of inconsequential print statements, designed to ``fool'' the models \cite{srikant2021generating,  yefet2020adversarial,gao2023discrete,allamanis2018learning}.  However none of the previous works study the vulnerability of the much larger LLMs used in coding tasks to the same adversarial perturbations.

Recent work~\cite{UniversalAttack23} describes a gradient-based attack to LLMs (but not specifically for coding tasks).  The goal of the attack is to find a suffix to potentially harmful user prompts, e.g., ``How to make a pipeline bomb'', so the combined prompt would break typical LLM alignment filters. The work shows the  adversarial suffixes found by the attack on open-source LLMs transfer well to commercial models like ChatGPT and Claude. In contrast we study LLMs specifically for coding tasks and show that attacks obtained on non-LLM models (such as seq2seq or code2seq) also transfer to LLMs. These attacks are much cheaper to compute, since the corresponding code models are much smaller than the LLMs.

 Classical defense approaches involve adversarial training, i.e., training using perturbed inputs \cite{madry2019deep,bielik2020adversarial,advtrainingcodeicse22,ramakrishnan2020semantic}. Such defenses remain challenging, as they often reduce model accuracy. To perform adversarial training for LLMs one can explore fine-tuning the pre-trained large models, as allowed by current APIs. However adversarial training for LLMs is likely very expensive and of limited effect. Recently, defenses based on either filtering the inputs or outputs of LLMs have been proposed but these approaches are geared towards defending the models against ``jailbreaking'' attacks, i.e., attacks that cause LLMs to produce toxic or harmful text~\cite{jain2023baseline,kumar2023certifying,robey2023smoothllm}. We instead propose cost-effective self-defense prompting techniques to counteract alterations caused by {\em known} adversarial code perturbations by leveraging an LLM’s own capabilities, through prompt instructions.

\section{Conclusion}
\label{sec:conclusion}
In this paper we have shown that adversarial examples obtained with a smaller code model are indeed transferable to modern LLMs, weakening LLMs' performance on coding tasks. We further proposed and evaluated novel prompt-based self-defense strategies that do not require retraining the LLMs but instead utilize LLMs' powerful capabilities to perform in-context learning and understand human instructions as well as code. We also presented meta-prompting, a technique that leverages LLMs themselves to synthesize the self-defense prompts. Our experiments show that prompt-based self-defense is effective, and meta-prompting can lead to even more effective prompts than the ones that are manually crafted. We believe that leveraging LLMs for self-defense and for synthesizing prompts is a generally applicable strategy that could also be useful for other reasoning tasks. As future work,
we also plan to investigate techniques that frame the search for finding the optimal self-defense prompts as an optimization problem solved either via gradient-based search over LLM parameters or via techniques suitable for black-box LLMs but requiring multiple calls to the LLM.

\bibliographystyle{ACM-Reference-Format}
\bibliography{references}

\end{document}